\documentclass[runningheads]{llncs}
\usepackage{algorithm,algpseudocode,amsmath,amsfonts,booktabs,xspace,graphicx,array,xcolor}
\usepackage[bb=dsserif]{mathalpha}

\makeatletter
\DeclareRobustCommand\onedot{\futurelet\@let@token\@onedot}
\def\@onedot{\ifx\@let@token.\else.\null\fi\xspace}

\def\eg{\emph{e.g}\onedot} 
\def\ie{\emph{i.e}\onedot}

\def\etal{\emph{et al}\onedot}
\makeatother

\newcolumntype{x}[1]{>{\centering\arraybackslash\hspace{0pt}}p{#1}}

\begin{document}
\title{Find the Cliffhanger: \\ Multi-Modal Trailerness in Soap Operas}
\author{
Carlo Bretti\inst{1}\orcidID{0000-0002-4489-5454} \and
Pascal Mettes\inst{1}\orcidID{0000-0001-9275-5942} \and
Hendrik Vincent Koops\inst{2}\orcidID{0000-0002-6980-7027} \and
Daan Odijk\inst{2}\orcidID{0000-0003-0369-8857} \and
Nanne van Noord\inst{1}\orcidID{0000-0002-5145-3603}}
\authorrunning{Bretti et al.}
\institute{Informatics Institute, University of Amsterdam, Netherlands\\
\email{\{c.bretti,p.s.m.mettes,n.j.e.vannoord\}@uva.nl}\\ \and
RTL Nederland, Netherlands\\
\email{\{vincent.koops,daan.odijk\}@rtl.nl}}
\maketitle              %
\begin{abstract}
Creating a trailer requires carefully picking out and piecing together brief enticing moments out of a longer video, making it a challenging and time-consuming task. This requires selecting moments based on both visual and dialogue information. We introduce a multi-modal method for predicting the trailerness to assist editors in selecting trailer-worthy moments from long-form videos.
We present results on a newly introduced soap opera dataset, demonstrating that predicting trailerness is a challenging task that benefits from multi-modal information. Code is available at \url{https://github.com/carlobretti/cliffhanger}

\end{abstract}

\section{Introduction}

Trailers are short previews of long-form videos with the aim of enticing viewers to watch a movie or TV-show~\cite{bordwellFilmArtIntroduction2020}. Professional editors create trailers by selecting and editing together moments from long-form video. Determining which moments to select can be a time-consuming task as it involves scanning the entire long-form video and picking moments that match aesthetic or semantic criteria. Those selected moments are considered to have high \textit{trailerness} - they represent the moments that are most suitable to be used in a trailer. 
By learning to automatically recognize moments that have high trailerness it becomes possible to support editors in creating trailers - as well as boost their creativity by selecting or recommending moments they may not have initially picked. 

Trailers are a key component of soap operas, as these follow a regimented format with daily episodes, fixed time slots, and every episode ending with a trailer enticing the viewer to stay tuned. Unlike trailers for movies, which may feature cinematographically spectacular footage, the mainstay for soap opera trailers is the continuation of prominent storylines, i.e., how the events from the current episode unfold in future episodes. As such, the context in which the material was created and the format followed may strongly influence what moments are used in trailers. This context for television programming, and soap operas in particular, due to their grounding in everyday situations, reflects national and cultural identities \cite{turnerCulturalIdentitySoap2005}, therefore requiring learning approaches that take into account both the complexity and the idiosyncrasies of the source material.

Crucially, what determines the trailerness of a moment may thus be conveyed through various modalities, e.g., a shot that is particularly visually attractive or exciting, or a line of dialogue that is particularly funny. While previous works have aimed to extract moments with high trailerness~\cite{wangLearningTrailerMoments2020}, this multi-modal aspect has not received sufficient attention. Next to multi-modality, we find that the time scale of trailer-worthy moments is understudied, with most approaches focusing only on shot-level moments. Yet, trailers are short in duration and typically fast-paced, therefore being able to select shorter-than-shot moments would be highly desirable.
To select moments with high trailerness most previous works on trailer generation either rely on dense annotations~\cite{smeatonAutomaticallySelectingShots2006} or external information that is integrated in a hand-crafted manner~\cite{xuTrailerGenerationPoint2015,smithHarnessingAugmentingCreativity2017,papalampidiFilmTrailerGeneration2021}. Instead, we propose to leverage existing trailers to learn what moments have high trailerness, based on both visual and subtitle information across multiple time-scales. 

Our results show that trailer generation is a highly challenging and subjective task and with our findings, we demonstrate the benefits of a multi-modal approach for trailer generation. In the following, we discuss related work, our proposed multi-scale and multi-modal method, our soap opera trailerness dataset, and present the findings in more detail.

\section{Related Work}
Movies and TV shows are rich and inherently multimodal sources of information. Within the domain of movie and TV series data, different tasks have been researched~\cite{everinghamHelloMyName2006,hanjalicAutomatedHighlevelMovie1999,tapaswiMovieQAUnderstandingStories2016,wuLongFormVideoUnderstanding2021}. Within this field, we specifically focus on trailer generation, to reduce a long-form video into a short-form video that could be used as a trailer~\cite{papalampidiFilmTrailerGeneration2021,smeatonAutomaticallySelectingShots2006,wangLearningTrailerMoments2020}. 
A closely related task to trailer generation and a common way to produce shorter videos from longer ones is video summarization~\cite{fajtlSummarizingVideosAttention2019,zhangVideoSummarizationLong2016}. In the following, we discuss works from both directions and how they differ.

\subsection{Video Summarization}

The aim of video summarization is to reduce the length of a long video in a brief and faithful manner. There have been both supervised and unsupervised approaches. Unsupervised methods often focus on generating a summary that is most representative of the original video in terms of its ability to reconstruct the original video~\cite{mahasseniUnsupervisedVideoSummarization2017}. Since our focus is on trailers, and scenes in a trailer are meant to elicit attention rather than provide a comprehensive summary, we focus our discussion of video summarization literature on supervised methods. One relevant line of work in supervised learning consists of modeling videos as sequences of frames~\cite{zhangVideoSummarizationLong2016}. Building on this, another avenue focused on first modeling frames within shots in a video as a sequence and then modeling entire videos as a sequence of shots~\cite{zhaoHierarchicalMultimodalTransformer2022,zhaoHierarchicalRecurrentNeural2017,zhaoHSARNNHierarchicalStructureAdaptive2018,zhaoTTHRNNTensorTrainHierarchical2021}. Various methods have also employed attention to model the relationship between multiple frames~\cite{fajtlSummarizingVideosAttention2019,jiDeepAttentiveSemantic2020,jiVideoSummarizationAttentionBased2020,liExploringGlobalDiverse2021,liuLearningHierarchicalSelfAttention2019}.
Lastly, multimodal summarization has been explored in a few cases. \cite{evangelopoulosMultimodalSaliencyFusion2013} provided a method to identify salient moments in a video using multimodal hand-crafted features. Some have additionally proposed query-dependent video summarization based on a textual query~\cite{huangGPT2MVSGenerativePretrained2021,huangQuerycontrollableVideoSummarization2020}. Combining deep features extracted from audio and video at multiple levels using attention has been explored recently~\cite{zhaoHierarchicalMultimodalTransformer2022}.

Within video summarization, the goal is to provide a faithful and brief representation of the source material. In contrast, we can characterize trailer generation as a form of biased video summarization aimed at enticing the audience to watch subsequent material. Trailer generation is therefore a more subjective task, an aspect which will be highlighted in the following section.

\subsection{Trailer Generation}

Trailer generation is not a well-studied topic, with few works focusing on it specifically.
Earlier approaches used hand-crafted features, whereas later works focus on learning-based approaches, primarily by incorporating external information. 

From the earlier works, the work in~\cite{smeatonAutomaticallySelectingShots2006} is most akin to the learning-based approaches in that a supervised model is trained on hand-crafted visual and audio features to classify shots as trailer worthy. This approach differs from other earlier works which used a more rule-based method, either informed by the typical structure of a Hollywood trailer~\cite{vonwenzlawowiczSemanticVideoAbstracting2012}, or by extracting key symbols of a movie (e.g., title logo and theme music)~\cite{irieAutomaticTrailerGeneration2010}.

Among more recent works, there is a wider emphasis on using a learning-based approach, where the majority of approaches rely on external information.
For example, in~\cite{xuTrailerGenerationPoint2015} a model of visual attractiveness is learned by leveraging eye-movement data, which is then used to generate a trailer tailored to a piece of music using a graph-based algorithm.
Similarly,~\cite{papalampidiFilmTrailerGeneration2021} uses narrative structure and sentiment in screenplays to generate trailers. 

In~\cite{smithHarnessingAugmentingCreativity2017} presented an approach that was used to create a trailer for the 2016 sci-fi film \textit{Morgan}. Their approach uses PCA on features extracted from different modalities (e.g. emotions, objects, scenes, and sounds) to select 3 principle components to be used as a scoring mechanism for suggesting shots to editors. Their final trailer is produced by editors who use the system as a selection mechanism.

In contrast to these works, our work is based on learning trailer-worthy sequences through ranking, which is more closely related to recent video summarization approaches. A similar approach to ranking shots rooted in visual learning was proposed in~\cite{wangLearningTrailerMoments2020}. Here, a model is trained to co-attend pairs of movies and trailers, and obtain correlation scores of shots across end-to-end. This can then be leveraged to learn a ranking model that ranks shots based on whether they are determined to be key moments of a movie. 
Our proposed method is distinct from previous works in that it consists of a multi-modal learning approach using visual and text inputs at multiple scales. Moreover, we leverage trailers of long videos to directly obtain trailerness annotations for training. In particular, the multi-modal and multi-scale aspects of trailer generation were understudied, whereas we find that these are greatly beneficial for performance.
\section{Method}
\begin{figure*}[t]
    \centering
    \includegraphics[width=0.9\textwidth]{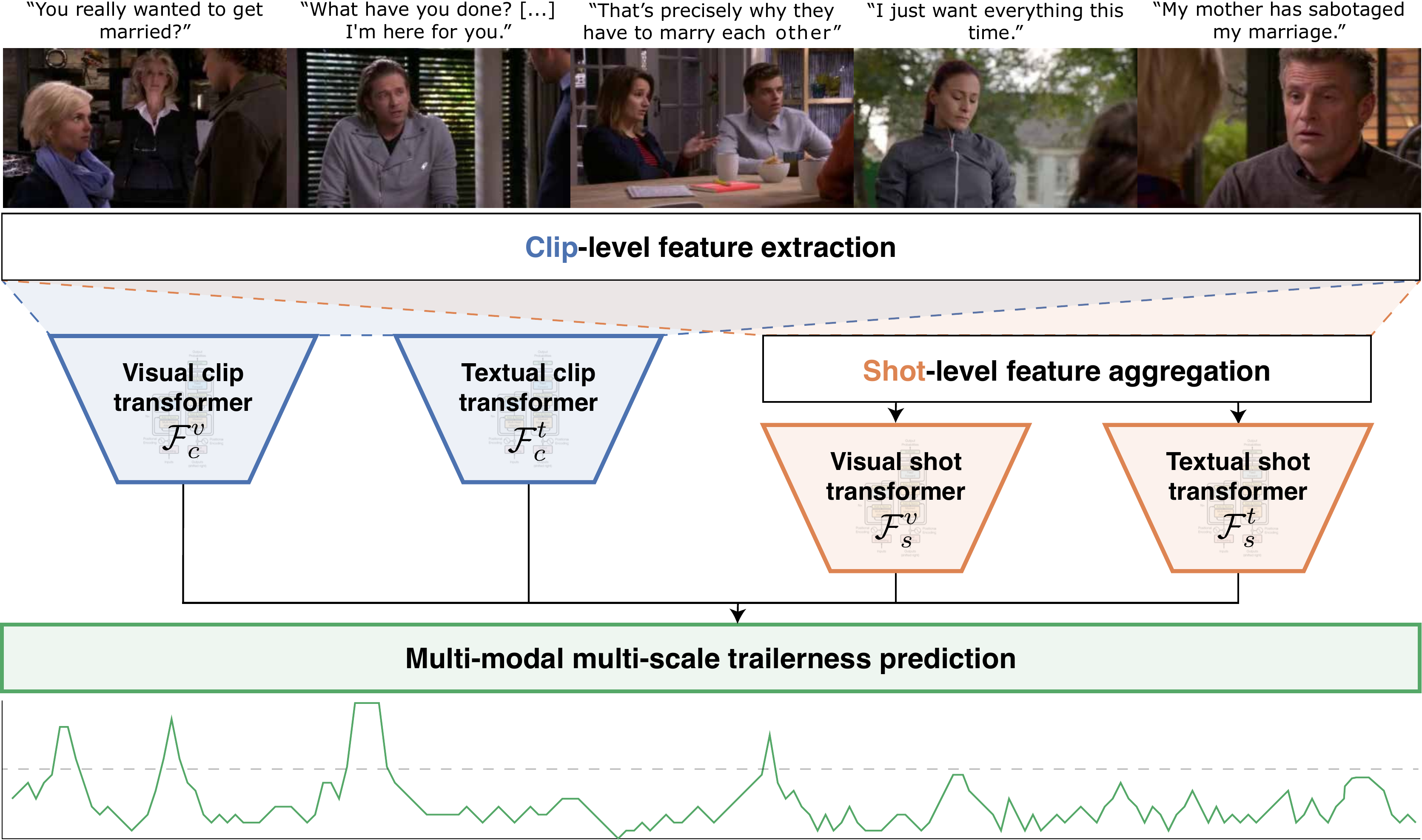}
    \caption{Estimating trailerness in videos with our Trailerness Transformer. Given a video denoting a movie or tv series episode, we first encode clip-level and shot-level encodings for both the visual and textual video modalities. We then train transformers for each combination of modality and temporal scale, after which we aggregate the trailerness predictions of all transformers.}
    \label{fig:methods}
\end{figure*}
In this section, we will first define a notion of trailerness, a method for obtaining trailer labels from editor selections, and finally a method for predicting trailerness scores as depicted in Figure~\ref{fig:methods}.

\subsection{The Trailerness of Video}

We are interested in discovering which parts of a TV episode or movie are most suitable to be used in a trailer. More generally, for all frames $x_f$ in a video $m$ we define \emph{trailerness} as a score that indicates how suitable each frame is to be selected for a trailer. Subsequently, sequences of high trailerness can be used by editors to piece together trailers.

Each video is denoted as $m = \{(v, t)\}$, where $v \in \mathbb{R}^{H \times W \times C \times |x_f|}$ denotes the visual track and $t_i = (f_{\text{start}}, f_{\text{end}}, b)$ denotes the natural language sentence $b$, and the start and end frame for the $i$th subtitle, both tracks contain complementary information for determining trailerness.

For learning trailerness, we are given a training collection of videos $M$ and a collection of corresponding trailers $G$. Each video corresponds to one trailer, hence $|M|=|G|$. Moreover, each trailer $g \in G$ solely relies on content from its corresponding video, \ie  $g \subset m$. The trailers are generated by professional editors to obtain high-quality selections. In order to predict which frames of a new video are worthy of being in a trailer, we first need to obtain frame-level annotations for all videos in $M$ by matching them visually to the editor-standard trailers. Then, we approach the problem as a localized scoring problem where we assign each frame $x_{f_j} \in x_f$ a trailerness score and obtain trailerness binary predictions, which can then be evaluated against the editor-standard labels.

\subsection{Trailer Labels from Editor Selections}
To establish a ground truth for trailerness, we perform visual matching per frame. We obtain annotations for all videos by matching each frame from videos $M$ against the trailers $G$.
Here, we outline a method to obtain frame-level annotations which we later aggregate to obtain clip-level labels for optimization.

First, we utilize a hashing function $\mathcal{F}_\text{hash}$ to compute a perceptual hash for each frame in video track $v$ for every video $m \in M$. We then do the same for all trailers $g \in G$.
For similarity search, we compute the Hamming distance between the hashes of frames in the trailers and the frames in the long videos. By doing so, we obtain a fast visual matching score between each trailer frame $r \in g$ and long video frames, which we then threshold to obtain editor-standard binary labels. For each individual frame $x_{f_j}$ in each video $m$, we first determine the following hashing distance:
\begin{equation}
y_{f_j} = [\![ \bigg( \min_{r \in g_i} d_h(\mathcal{F}_\text{hash}(v_{f_j}), \mathcal{F}_\text{hash}(r)) \bigg) < \tau ]\!],
\end{equation}
with $d_h(\cdot, \cdot)$ the Hamming distance and $\tau$ a hashing distance threshold, stating the minimum required similarity between video and trailer frames. Stacking the labels of all frames in video $m$ results in a label vector $y_f \in \{0, 1\}_{j=1}^{|x_f|}$ indicating which frames were deemed trailer-worthy by a professional editor.

\subsection{Multi-scale and Multi-modal Trailerness Transformer} 

Given dataset $M$ where each video $m = \{(v, t, y_f)\}$ contains video frames, subtitles, and annotations, we seek to obtain an end-to-end model that provides trailerness predictions for all frames in a test video. We hypothesize that such a prediction is best made when considering both visual and textual modalities over both long- and short-term intervals. To that end, we introduce a Trailerness Transformer that performs multi-modal and multi-scale trailerness prediction. Below we outline the three main stages of our approach, namely encoding, multi-modal and -scale transformers, and prediction aggregation.
\\\\
\noindent\textbf{Visual and Textual Encoding.}
We operate on four types of sequences, varying in temporal scale and modality. We consider a short clip-level temporal scale of 64 frames (2.56 seconds at 25 fps) and a longer shot-level temporal scale, with each shot consisting of multiple clips.
For the visual modality, we employ a pre-trained video embedding function $\mathcal{F}_V(\cdot)$ to extract features for each 64-frame clip from a~\cite{carreiraQuoVadisAction2017}. A video $v$ is divided into 64-frame clips and is given as $v_c=\{v_{c_j}\}_{j=1}^{|v_c|}$. Then, for each clip we extract features as $E_{c_j}^v = \mathcal{F}_V(v_{c_j})$, yielding $E_c^v=\{E_{c_j}^v\}_{j=1}^{|v_c|}$ for a long video $m$.
For the textual stream, we concatenate any subtitle with temporal overlap with the clip timeframe for each clip. We then take the corresponding features from a pre-trained language model, $E_c^t = \{\mathcal{F}_T(t_{c_j})\}_{j=1}^{|t_c|}$. 

Besides a local clip-level view, we also investigate a longer shot-level scale.
We define shots as video sequences for which the boundaries are represented by shot transitions. These shot boundaries are determined by a shot transition detector $\mathcal{F}_\text{shots}$ which outputs cut probabilities for each frame~\cite{soucekTransNetV2Effective2020}. By using shots, we have more coarse sequences that naturally follow from the source material.
For visual features at a shot level, we aggregate over the extracted features $E_c^v$ based on the shot boundaries obtained through $\mathcal{F}_\text{shots}$ to then obtain shot-level visual features $E_s^v$. On the other hand, for the textual features, we again first concatenate subtitles as for clips, this time based on the overlap between subtitle boundaries and shot boundaries to obtain $t_s$. Shot-level textual features can then be computed directly using the pre-trained text embedding function $\mathcal{F}_T$, with $E_s^t = \{\mathcal{F}_T(t_{s_j})\}_{j=1}^{|t_s|}$. Finally, we obtain four streams of features, depending on the combination of  scale and modality, $E_c^v, E_c^t, E_s^v, E_s^t$ for a long video $m$. We will use $E_\ast$ to refer to multiple scales and $E^{\star}$ to refer to multiple modalities.
\\\\
\noindent\textbf{Modality- and Scale-specific Transformers.}
On the four combinations of modality and temporal scale, we train an individual transformer, where all transformers follow the architecture of Vaswani \etal~\cite{vaswaniAttentionAllYou2017}.
We conceptualize our problem as a sequence-to-sequence binary classification task, where given a video $m$ as input features $E_\ast^\star \in \mathbb{R}^{D \times |x_\ast|}$. $D$ is the number of dimensions in output from video feature extractor $\mathcal{F}_V(\cdot)$ or $\mathcal{F}_T(\cdot)$ for text, while $|x_\ast|$ the  number of sequences for a given video $m$ depending on the scale at play.

We consider a transformer architecture $\mathcal{F}_\ast^\star(\cdot)$ consisting of a linear layer, positional encodings, two transformer encoder blocks with hidden dimensionality $d_k$, and finally, an output linear layer followed by a sigmoid to obtain trailerness scores $O_\ast^\star$. We first map the input features through a linear layer and add the positional encodings~\cite{vaswaniAttentionAllYou2017} as given by 
\begin{equation}
PE_{(\text{pos},h)}
\begin{cases}
    \sin \Bigl(\frac{\text{pos}}{10000^{h/d_k}}\Bigr) & \text{ if }h\text{ mod } 2 = 0 \\
    \cos \Bigl(\frac{\text{pos}}{10000^{(h-1)/d_k}}\Bigr) & \text{ otherwise}
\end{cases}
\label{eq:pe}
\end{equation}
where pos represents the temporal position of a token (in our case either a clip or a shot) within the full video sequence and $h$ represents the specific dimension.
The resulting output is then fed through the transformer encoder blocks.
Each transformer encoder block is composed of two sub-layers: a multi-head attention mechanism (MHSA) and a multi-layer perceptron (MLP). A residual connection~\cite{heDeepResidualLearning2016} is added to the output of each sublayer, and the resulting output undergoes layer normalization~\cite{baLayerNormalization2016}. 

The transformer encoder output is then fed through a final linear layer followed by a sigmoid, to obtain trailerness predictions $O_\ast^\star = \mathcal{F}_\ast^\star(E_\ast^\star)$ with $O_\ast^\star \in \mathbb{R}^{1 \times |x_\ast|}$. 
The trailerness labels are provided at frame level, but we investigate clip- and shot-level encodings in our approach. A clip is deemed positive if at least one third of the frames within the clip are positive. Similarly, a shot is deemed positive if at least one third of the clips contained within the shot are positive. For a video $m$ this results in clip-level annotations $y_c \in \{0, 1\}_{j=1}^{|x_c|}$ and shot-level annotations $y_s \in \{0, 1\}_{j=1}^{|x_s|}$.
Focal loss~\cite{linFocalLossDense2017} has been found to reliably work well for highly imbalanced data and is therefore used for training:
\begin{equation}
\mathcal{L}= - \alpha (1 - o_\textrm{p})^\gamma \log (o_\textrm{p}),
\label{eq:focal}
\end{equation}
with 
\begin{equation}
o_\textrm{p}=
\begin{cases} 
o_{\ast_j}^* &\text{if $y_{\ast_j} = 1$},\\ 
1 - o_{\ast_j}^* &\text{otherwise}
\end{cases}
\label{eq:pt}
\end{equation}
where $o_{\ast_j}^\star \in O_\ast^\star$ is a single prediction for a subsequence for a video $m$. The parameter $\alpha$ controls the balance between positive and negative examples, emphasizing positive examples in this case. On the other hand, $\gamma$ determines a modulating factor that allows the loss to distinguish between easy and hard examples.
\\\\
\textbf{Fusing Transformer Predictions.}
We obtain four different trailerness prediction streams for a video, one for each combination of scale and modality.
We seek to fuse these predictions to make use of the complementary information across the four transformers. However, the number of sequences $|x_\ast|$ depends on the size of the sequences, \ie whether they are clips or shots.
At test time, we upsample all predictions from each stream from either clip- or shot-level predictions to frame-level predictions. This allows us to fairly compare outputs from different streams. We can also then perform late fusion of predictions using different possible combinations of the four multi-modal multi-scale streams. This can be done simply by averaging frame-level prediction likelihoods for the considered streams~\cite{snoekEarlyLateFusion2005}.
\begin{table*}[ht]
    \caption{Comparing our GTST trailer dataset to popular video summarization datasets and trailer moment detection datasets. * indicates the subset of VISIOCITY with only tv shows. Our dataset is unique in its combination of editor-standard labels, multi-modal nature, and availability for open research.}
    \centering
    \begin{tabular}{x{2.5cm}x{1.3cm}x{1.5cm}x{1.5cm}x{2.2cm}x{1.3cm}x{1.3cm}}
    \toprule
    Dataset                                                  & \#Videos & Avg. length & Goal          & Label type      & Multimodal & Openly available \\
    \midrule
    SumME~\cite{gygliCreatingSummariesUser2014}              & 25        & 2 min       & summary & user-generated  &            & \checkmark \\
    TVSum~\cite{songTVSumSummarizingWeb2015}                 & 50        & 4 min       & summary & user-generated  &            & \checkmark \\
    VISIOCITY~\cite{kaushalRealisticVideoSummarization2020}  & 67        & 55 min      & summary & user-generated  &            & \checkmark \\
    VISIOCITY*~\cite{kaushalRealisticVideoSummarization2020} & 12        & 22 min      & summary & user-generated  &            & \checkmark \\
    LSMTD~\cite{huangTrailersStorylinesEfficient2018}        & 508       & N/A         & trailers      & paired data     &            &            \\
    MovieNet~\cite{huangMovieNetHolisticDataset2020}         & 1100      & N/A         & trailers      & paired data     & \checkmark &            \\
    TMDD~\cite{wangLearningTrailerMoments2020}               & 150       & N/A         & trailers      & paired data     &            &            \\
    GTST               & 63        & 22 min      & trailers      & editor-standard & \checkmark & \checkmark \\
    \bottomrule \\
    \end{tabular}
    \label{tab:datasetcompare}
\end{table*}
\section{The GTST Dataset} \label{dataset}
To investigate multi-modal trailerness prediction we introduce the \textbf{GTST} dataset. This multi-modal dataset consists of 63 episodes from the long-running Dutch soap opera "Goede Tijden, Slechte Tijden". Each episode is around 20 minutes in length and for each episode, we have the visual video track and the time-coded subtitles.
A typical episode of GTST consists of three blocks: a recap sequence, the body of the episode, and a preview sequence. The recap sequence serves as a way to bring viewers up to speed with the plot. It summarizes the main plot points of relevant earlier episodes, with an emphasis on the preceding episode. On the other hand, the preview sequence serves as a way to persuade viewers to tune in for the next episode. It features clips from the next episode that are meant to catch viewers' attention, \ie cliffhangers or shocking revelations or events. In the context of soap operas, a preview of an episode represents a very close match to what a trailer represents for a feature film.
We will therefore rely on previews to obtain editor-standard trailerness labels for the original long videos, \ie episodes in the context of a soap opera.

Our dataset features long videos from a TV show and features readily available high-quality editor-standard labels compared to user-generated summaries~\cite{kaushalRealisticVideoSummarization2020,wangLearningTrailerMoments2020}. Table~\ref{tab:datasetcompare} compares different datasets for summarization and trailer generation to our dataset. All datasets for summarization are publicly available and they are densely labeled with user-generated annotations, but consist of only video data. The availability of data for trailer generation is much more limited, most datasets consist of commercial movies paired with their trailers. For these datasets no annotations are available, but the trailers could be used for supervision - as we demonstrate with our approach for extracting editor-standard pseudo-labels.
Overall, this table highlights the unique trailer-oriented, editor-standard, and multi-modal properties of GTST.
The dataset is divided into training/validation/test sets with 60\%/20\%/20\% per set respectively.
\section{Experiments}

We focus on two types of empirical evaluations: (i) quantitative evaluations of the trailerness transformer compared to editor selections, (ii) qualitative analyses of discovered trailerness.

\subsection{Setup}
\textbf{Encoding}.
For our experiments, we use three pre-trained models to obtain encodings and shot boundaries.  Visual features are extracted using I3D~\cite{carreiraQuoVadisAction2017} as video feature extractor $\mathcal{F}_V$, as provided in~\cite{iashinVideoFeatures2023}. The textual features are computed using a multilingual sentence embedding model $\mathcal{F}_S$~\cite{reimersSentenceBERTSentenceEmbeddings2019,reimersMakingMonolingualSentence2020}. This is a multilingual version of MiniLM~\cite{wangMiniLMDeepSelfAttention2020}, a compressed version of a pre-trained transformer. Lastly, TransNetV2~\cite{soucekTransNetV2Effective2020} serves as our shot boundary detector $\mathcal{F}_\text{shots}$.
As for the transformer architecture employed, we empirically set the number of attention heads to 4, and we use one transformer encoder block. For the focal loss, we empirically set the parameters $\alpha=0.95$ and $\gamma=1$ through validation.
\\\\
\noindent\textbf{Obtaining Trailer Labels}.
To obtain the per-frame binary trailerness labels for all videos in our dataset,
we first split the videos to isolate the body of the episode based on their known structure. We also extract the preview sequence from each episode that corresponds to the next episode. We then remove all the redundant sequences such as opening titles or title cards before and after ad breaks. After having extracted all the frames from the body of an episode and from its corresponding preview we compute perceptual hashes using Imagehash~\cite{buchnerImagehashPythonPerceptual2021}. We perform similarity search using FAISS~\cite{johnsonBillionScaleSimilaritySearch2021} and match the frames in a preview sequence to the original frames in the episode to obtain editor-standard trailerness labels.
\\\\
\noindent\textbf{Evaluation}.
For evaluation, we binarize predictions based on a threshold ($0.5$) and compute F1, precision, and recall between the frames predicted to be in a trailer and the frames selected by editors. We run the models with 5 different random seeds and report mean and standard deviation estimates for each metric.

\begin{table}[t]
    \caption{Trailerness prediction on individual transformer streams, in percentage \%. Overall, clip-level transformers obtain better F1 scores. Across modalities, visual clip-level trailerness boosts precision, while textual shot-level trailerness boosts recall, highlighting their complementary nature.}
    \centering
        \begin{tabular}{x{1.5cm}x{1.5cm}x{1.5cm}x{1.5cm}x{1.5cm}x{1.5cm}x{1.5cm}x{1.5cm}}
        \toprule
        \multicolumn{2}{c}{Clip-level} & \multicolumn{2}{c}{Shot-level} & & & \\
         Visual &  Textual &  Visual &  Textual &  F1 &  Prec. &  Rec. \\
        \midrule
            \checkmark &            &            &            & 6.9 {\scriptsize $\pm$ 1.0}  & 4.0 {\scriptsize $\pm$ 0.7}  & 28.0 {\scriptsize $\pm$ 11.8}  \\
                       & \checkmark &            &            & 5.5 {\scriptsize $\pm$ 1.3}  & 3.3 {\scriptsize $\pm$ 0.9}  & 18.2 {\scriptsize $\pm$ 2.8}  \\
                       &            & \checkmark &            & 5.0 {\scriptsize $\pm$ 2.6}  & 3.0 {\scriptsize $\pm$ 1.4}  & 26.2 {\scriptsize $\pm$ 25.0}  \\
                       &            &            & \checkmark & 5.2 {\scriptsize $\pm$ 0.5}  & 2.8 {\scriptsize $\pm$ 0.3}  & 40.9 {\scriptsize $\pm$ 11.6}  \\
        \bottomrule
        \end{tabular}
    \label{tab:individualstreams}
\end{table}

\subsection{Evaluating Modalities and Temporal Scales}

We first evaluate the four individual combinations of modalities and temporal scales for their ability to predict trailerness in unseen test videos.
In Table~\ref{tab:individualstreams}, we show the performance of our transformer-based approach across the four combinations.  We find that the visual clips stream $\mathcal{F}_c^v$ provides the best performance in terms of F1-score, (6.9\%). 
Moreover, the streams using features extracted at a clip level, \ie $\mathcal{F}_c^v$ (6.9\%) and $\mathcal{F}_c^t$ (5.5\%), perform better overall than the streams using features at a shot level $\mathcal{F}_s^v$ (5.0\%), $\mathcal{F}_s^t$ (5.2\%).

Only using the visual stream at a shot level, as is done in the trailer-based approach of Wang \etal~\cite{wangLearningTrailerMoments2020} leads to lower performance. Within the visual modality, a more localized approach seems to improve performance.

\subsection{Combining Modalities and Temporal Scales}
\begin{table}[t]
    \caption{Results for late fusion of different trailerness streams, in percentage \%. In our results, a triplet, combining modalities at a shot-level with clip-level visual predictions balances precision and recall best, as indicated by the F1 score.}
    \centering
        \begin{tabular}{x{1.5cm}x{1.5cm}x{1.5cm}x{1.5cm}x{1.5cm}x{1.5cm}x{1.5cm}}
        \toprule
        \multicolumn{2}{c}{Clip-level} & \multicolumn{2}{c}{Shot-level} & & & \\
         Visual &  Textual &  Visual &  Textual &  F1 &  Prec. &  Rec. \\
        \midrule
        \checkmark & \checkmark &            &            & 6.5 {\scriptsize $\pm$ 2.0}  & 3.9 {\scriptsize $\pm$ 1.1}  & 20.5 {\scriptsize $\pm$ 8.1}  \\
                   &            & \checkmark & \checkmark & 7.1 {\scriptsize $\pm$ 0.6}  & 4.0 {\scriptsize $\pm$ 0.3}  & 37.5 {\scriptsize $\pm$ 10.2}  \\
        \checkmark &            & \checkmark &            & 7.8 {\scriptsize $\pm$ 0.4}  & 4.7 {\scriptsize $\pm$ 0.2}  & 23.4 {\scriptsize $\pm$ 5.7}  \\
                   & \checkmark &            & \checkmark & 7.2 {\scriptsize $\pm$ 1.0}  & 4.2 {\scriptsize $\pm$ 0.5}  & 26.1 {\scriptsize $\pm$ 7.7}  \\
        \checkmark &            &            & \checkmark & 7.9 {\scriptsize $\pm$ 1.4}  & 4.7 {\scriptsize $\pm$ 1.0}  & 30.6 {\scriptsize $\pm$ 12.1}  \\
                   & \checkmark & \checkmark &            & 6.7 {\scriptsize $\pm$ 2.1}  & 4.5 {\scriptsize $\pm$ 1.3}  & 17.0 {\scriptsize $\pm$ 9.6}  \\
        \midrule
        \checkmark & \checkmark & \checkmark &            & 7.3 {\scriptsize $\pm$ 1.4}  & 4.6 {\scriptsize $\pm$ 0.8}  & 18.5 {\scriptsize $\pm$ 5.7}  \\
        \checkmark & \checkmark &            & \checkmark & 6.9 {\scriptsize $\pm$ 2.2}  & 4.2 {\scriptsize $\pm$ 1.2}  & 22.4 {\scriptsize $\pm$ 11.0}  \\
        \checkmark &            & \checkmark & \checkmark & 9.2 {\scriptsize $\pm$ 0.9}  & 5.6 {\scriptsize $\pm$ 0.4}  & 30.1 {\scriptsize $\pm$ 9.7}  \\
                   & \checkmark & \checkmark & \checkmark & 8.4 {\scriptsize $\pm$ 1.1}  & 5.2 {\scriptsize $\pm$ 0.8}  & 24.6 {\scriptsize $\pm$ 2.4}  \\
        \midrule
        \checkmark & \checkmark & \checkmark & \checkmark & 8.5 {\scriptsize $\pm$ 1.4}  & 5.3 {\scriptsize $\pm$ 0.8}  & 22.5 {\scriptsize $\pm$ 6.0}  \\
        \bottomrule
        \end{tabular}
    \label{tab:latefusion}
\end{table}

Through late fusion, we can consider different combinations of the four streams and their effect on predicting trailerness in test videos as we show inTable~\ref{tab:latefusion}. 
First, the top two rows in Table~\ref{tab:latefusion} showcase that a multi-modal fusion of predictions at a shot level (7.1\%) performs better than multimodal fusion at a clip level (6.5\%) in terms of F1. The third and fourth rows indicate that fusing clip-level and shot-level predictions for a single modality results in slightly better performance for the visual stream, with 7.8\% for visual clips and shots together against 7.2\% for predictions based on text fused at a clip and shot level.
Overall, multi-modal or multi-scale fusion boosts the performance of the weaker individual streams, with fusing the modalities at a shot level performing the best.

In rows five and six, we consider two pairs with neither matching modality nor matching size. Combining the visual feature at the clip level with text features at the shot level results in the highest F1 score for fusion based on two streams. 
Fusing clip-level predictions from text with shot-level predictions from visual features also boosts performance compared to the two individual streams as reported in Table~\ref{tab:individualstreams}, albeit not as strongly.  
Fusing across temporal scales and modalities results generally also results in an increase in performance, indicating that for combinations of two streams this is similarly beneficial.

The most striking result for triplets of streams is that by combining predictions from both modalities at a clip level with predictions from text at a shot-level we achieve our best performance in terms of F1, at 9.2\%, above the best performing individual stream (visual clips) at 6.9\%. 
Lastly, we show that a fusion of predictions from all streams does not result in an increase in performance and instead leads to a 0.7\% decrease in performance from our best model, from 9.2\% to 8.5\%.

\begin{figure}[t]
    \centering
    \includegraphics[width=0.65\columnwidth]{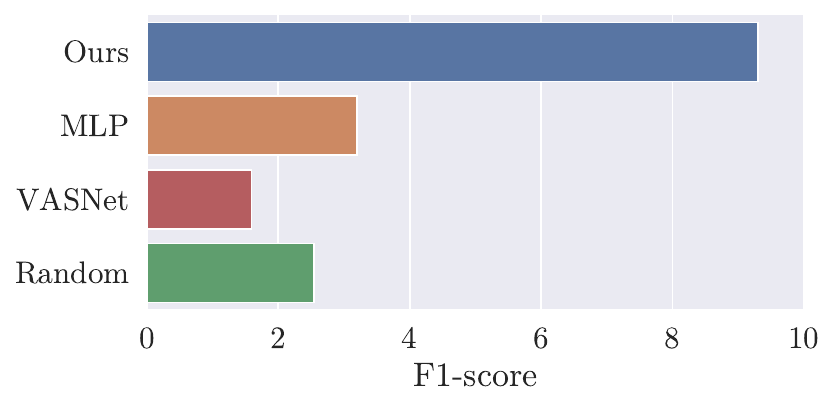}
    \caption{Baseline comparisons. An MLP-based architecture outperforms the random baseline and a frame-based summarization method (VASNet~\cite{fajtlSummarizingVideosAttention2019}), and our model outperforms them all by incorporating sequential order and temporal positioning.}
    \label{fig:comparisons}
\end{figure}
\subsection{Comparisons to Baselines}
In Figure~\ref{fig:comparisons} we compare results from our best-performing model to three baselines: random, a multi-layer perceptron (MLP), and a frame-based video summarization baseline VASNet \cite{fajtlSummarizingVideosAttention2019}. We show that our approach performs far better than a random baseline, where clips or shots are randomly assigned a trailerness score. For the second baseline, we train an MLP and separately perform hyperparameter tuning to find the best values for $\alpha$ and $\gamma$ in the focal loss
and we set $\alpha=0.98$ and $\gamma=1$. We additionally include early stopping as a regularizing factor. For the MLP, we find that the best results are given by fusing the clip and shot-level visual streams. We show that while an MLP architecture obtains results better than random, our choice of architecture showcases a 2.9x increase in F1, from 3.2\% to 9.2\%.  This is because our transformer-based approach incorporates sequentiality and positioning in its architecture, whereas the MLP baseline treats each subsequence individually, with no knowledge of sequentiality. 
As denoted by F1, our model provides a better overall performance, making it more suitable for predicting trailerness without overpredicting the positive class.
For our third baseline, we separately train VASNet, and tune hyperparameters. For a fair comparison we replace the mean squared error term with the focal loss, and follow the same evaluation procedure as ours. Hyperparameter tuning resulted in setting $\alpha=0.999$ and $\gamma=1$. We additionally adapt by splitting up long videos into shorter clips, to avoid complexity issues due to the self-attention operation over too long of a sequence. VASNet results in an F1 of 1.6\% below the performance of the other baselines and our model. This highlights the difference between summarisation and trailer generation, despite their similar premise.

\subsection{Qualitative Results}

\begin{figure*}
    \centering
    \includegraphics[width=\textwidth]{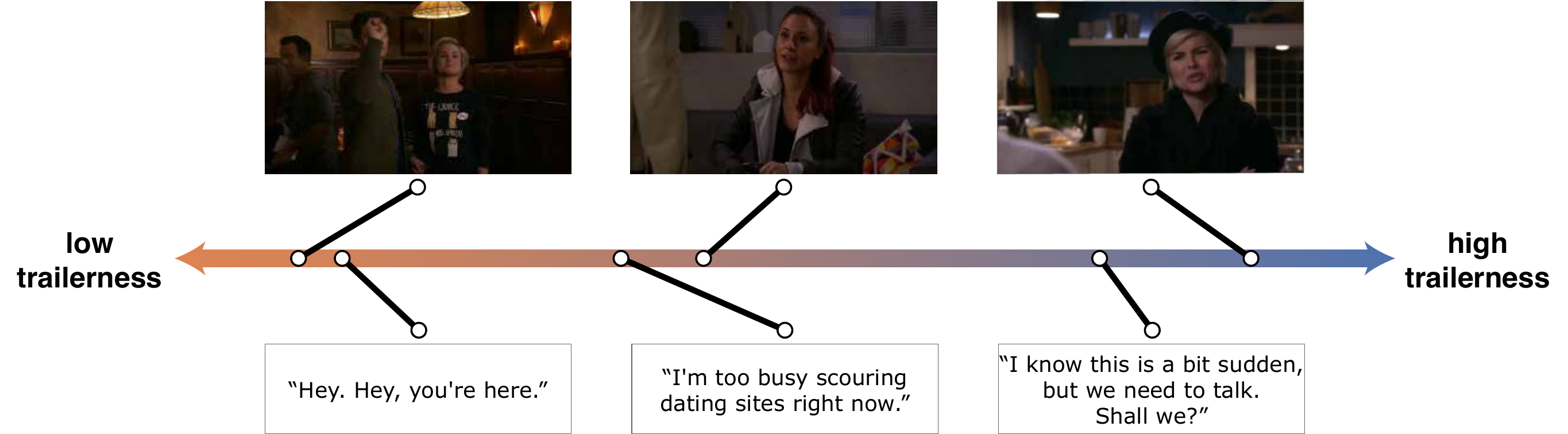}
    \caption{Qualitative results for visual and text streams at a clip level individually. Emotionally-charged visuals and urgent calls to action in text yield higher trailerness than transitory visuals and playful subtitles.}
    \label{fig:qualitative-individual}
\end{figure*}

\begin{figure*}
    \centering
    \includegraphics[width=\textwidth]{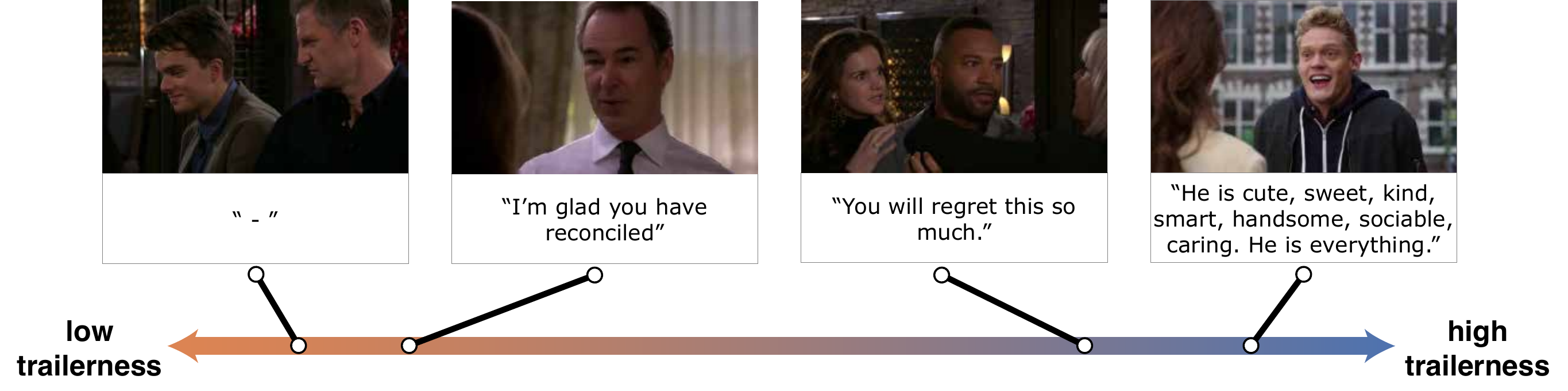}
    \caption{Qualitative results for our best-performing model. Scenes with bright visuals and emphatic dialogue yield higher trailerness than scenes with generic visuals and a lack of dialogue.}
    \label{fig:qualitative-best}
\end{figure*}

To gain more insight into our approach, we show qualitative results for different models in Figures~\ref{fig:qualitative-individual} and~\ref{fig:qualitative-best}. In Figure~\ref{fig:qualitative-individual} we relate textual and visual information to trailerness independently, showing how emotionally charged visuals and urgent textual calls to action yield higher trailerness in unimodal settings.

In Figure~\ref{fig:qualitative-best} we relate trailerness to the results of the best performing multi-stream model highlighting the interplay between modalities.

\section{Conclusion}

Which moments in a long-form video are suited for a trailer depends on a variety of factors, including the creative style of the editor and narrative aspects (\eg avoiding spoilers). This subjective nature makes selecting moments with high trailerness in a fully automatic manner a challenging task.  
We presented an approach that leverages existing trailers to generate annotations of trailerness, and use this to train a multi-modal and multi-scale model to predict trailerness underscoring the complexity of trailer generation and how it benefits from contextual information.

\subsubsection*{Acknowledgements}
This research was made possible by the TKI ClickNL grant for the AI4FILM project.

\bibliographystyle{splncs04}
\bibliography{zotero_bbt.bib}

\end{document}